\documentclass{article}

\usepackage[final,nonatbib]{nips_2016}

\usepackage[utf8]{inputenc} 
\usepackage[T1]{fontenc}    
\usepackage{url}            
\usepackage{booktabs}       
\usepackage{amsfonts}       
\usepackage{nicefrac}       
\usepackage{microtype}      

\usepackage{multirow}
\usepackage{threeparttable}
\usepackage{graphicx}
\usepackage{adjustbox}
\usepackage{subcaption}
\usepackage{mathrsfs}
\usepackage{amsmath}
\DeclareMathOperator*{\argmin}{argmin}
\usepackage{pgfplots}
\usepackage{lettrine}

\usepackage{pgfplotstable}
\usepackage{tikz}
\usetikzlibrary{patterns}

\usepackage{breakurl}
\usepackage[breaklinks]{hyperref}

\newcommand{\specialcellc}[2][c]{\begin{tabular}[#1]{@{}c@{}}#2\end{tabular}}

\title{Spiking Deep Residual Network}

\author{
  Yangfan Hu\\
  Zhejiang University\\
  \texttt{huyangfan@zju.edu.cn} \\
  \And
  Huajin Tang\\
  Zhejiang University\\
  \texttt{htang@zju.edu.cn} \\
  \And
  Gang Pan\\
  Zhejiang University\\
  \texttt{gpan@zju.edu.cn} \\
}

\begin{document}
\maketitle
\begin{abstract}
     Spiking neural networks (SNNs) have received significant attention for their biological plausibility. SNNs theoretically have at least the same computational power as traditional artificial neural networks (ANNs). They possess potential of achieving energy-efficiency while keeping comparable performance to deep neural networks (DNNs). However, it is still a big challenge to train a very deep SNN. In this paper, we propose an efficient approach to build a spiking version of deep residual network (ResNet). ResNet is considered as a kind of the state-of-the-art convolutional neural networks (CNNs). We employ the idea of converting a trained ResNet to a network of spiking neurons, named Spiking ResNet (S-ResNet). We propose a shortcut conversion model to appropriately scale continuous-valued activations to match firing rates in SNN, and a compensation mechanism to reduce the error caused by discretisation. Experimental results demonstrate that, compared with the state-of-the-art SNN approaches, the proposed Spiking ResNet achieves the best performance on CIFAR-10, CIFAR-100, and ImageNet 2012. Our work is the first time to build a SNN deeper than 40, with comparable performance to ANNs on a large-scale dataset.
\end{abstract}

\section{Introduction}
\lettrine[lines=2]{S}{piking} neural network (SNN) is considered as a promising approach to build an intelligent machine as efficient as the  human brain. Compared with conventional ANN, SNN is more biologically realistic and its neurons communicate with each other via discrete events (spikes) instead of continuous-valued activations. Theoretically, SNN can approximate any function as ANN \cite{maass1997networks}. Such system is updated asynchronously as event arrives, thus reduces number of operations required at each time step. Recent developments show that SNNs can be emulated by neuromorphic hardware such as TrueNorth \cite{merolla2014million}, SpiNNaker \cite{furber2014spinnaker} and ROLLS \cite{ning2015reconfigurable} with several orders of magnitude less energy consumption than that by contemporary computing hardware. Furthermore, SNNs are inherently suitable for processing data from emerging AER-based (address-event-representation) sensors that have low redundancy, low latency and high dynamic range, such as dynamic vision sensor (DVS) \cite{lichtsteiner2008128}. Visual systems \cite{orchard2015hfirst,amir2017low} constructed with SNN, DVS and neuromorphic processors have demonstrated their capacity in solving visual tasks as well as prominent energy-efficiency.

At current stage, how to train a SNN remains an open challenge due to the discontinuity of its spiking mechanism. Many efforts have been made to solve this problem. These methods can be categorised into four classes. The first class seeks to make SNN differentiable with some approximations and apply gradient descent. The second class takes inspiration from biological neurons and utilises synaptic plasticity rules. The third class views SNN as a stochastic process and learns with probabilistic inference. However, the three kinds of approaches are not yet well-developed and cannot deal with deep architectures. The last kind of approaches employ the idea of conversion to narrow the gap of performance between SNN and ANN. It trains a conventional ANN and builds a conversion algorithm to map the weights to an equivalent SNN. Although this approach can build deeper SNNs than the first three approaches, it is still a big challenge to build a large-scale SNN, for the performance after conversion will drop rapidly when the network architecture goes deeper, e.g. more than 30 layers.

In this paper, we investigate the learning of deep SNN by converting residual network \cite{he2016deep}, a cutting-edge CNN architecture that allows a network to go extremely deep and has achieved great success in many applications, to its spiking equivalent. Previous conversion methods are not applicable to residual network for the structural difference between residual structure and conventional linear structure. We design a shortcut conversion model to jointly normalise synaptic weights in spiking residual network. Furthermore, we find that accumulating propagation error is critical to obstruct the loss-less conversion of large-scale network and develop an effective method to compensate propagation error by slightly saturating firing rates of neurons and forcing neurons at deeper layers to response more quickly. Our method achieves the state-of-the-art performance with MNIST, CIFAR-10, CIFAR-100 and ImageNet. The preliminary work of this paper was released on the arxiv 2018, April 28 \cite{hu2018spiking}.

\section{Related Work}
In SNN, information is propagated by discrete events, i.e., spikes. Therefore, the indifferentiability makes classic learning methods for ANN, such as backpropagation and its variants, incompatible with SNN. A couple of learning algorithms have been proposed to build SNN, which can be categorised as follows:

1. Gradient descent based algorithms. This approach aims to overcome non-linearity in SNN and apply gradient descent optimisation algorithms. Early work goes back to $Spikeprop$ \cite{bohte2002error}, which implemented backpropagation in SNN by assuming the potential function to be differentiable for a small region around firing time. In a way similar to backpropagation, $tempotron$ learning rule \cite{gutig2006tempotron} was derived from a cost function defined by membrane potential and spike timing. More recently, direct training of SNN with stochastic gradient descent (SGD) was demonstrated in \cite{lee2016training,neftci2017event,mostafa2018supervised,shrestha2018slayer,zenke2018superspike,huh2018gradient,wu2019direct}.

2. Learning with synaptic plasticity rules. This approach utilises synaptic plasticity rules such as spike-timing-dependent plasticity (STDP) \cite{bi1998synaptic}. Evidenced by plentiful neurophysiological data, STDP has attracted the interest of many researchers. In \cite{masquelier2007unsupervised,masquelier2010learning}, STDP has been proven able to select visual features in an unsupervised manner. Stable learning over time was exhibited in \cite{diehl2015unsupervised} with STDP rules, lateral inhibition and homoeostasis employed. In \cite{liu2018event}, event-driven continuous STDP was proposed to recognise patterns from time-coded spike train. Rathi et al. \cite{rathi2018stdp} demonstrated the efficiency in energy and area with a pruned SNN and self-taught STDP learning rule.

3. Statistical algorithms. Although most algorithms treat SNN as a deterministic system, evidence suggests that neural network in brain resembles stochastic machine \cite{berkes2011spontaneous}. In \cite{maass2014noise}, noise and uncertainty are considered as beneficial factors that facilitate statistical learning and self-organisation in SNN. In \cite{huang2016bayesian}, a SNN has been shown to perform Bayesian inference, coinciding with the suggestion that Bayesian inference is prevalent in cognition behaviours. In \cite{neftci2016stochastic}, a network model with stochastic synapses was presented for Monte Carlo sampling and unsupervised learning.

4. Conversion methods. The study of converting a pre-trained ANN to its equivalent SNN began as Perez-Carrasco et al. \cite{perez2013mapping} introduced an approach to obtain the weights of SNN by scaling the weights of its CNN counterpart in accordance with the parameters of Leaky Integrate-and-Fire (LIF) \cite{gerstner2002spiking} spiking neuron.

One kind of conversion algorithms builds the mapping between ANN and SNN by fiddling the activation function/artificial neuron to approximate the average firing rate of spiking neurons. Methods in this category include training and mapping with Siegert neuron \cite{oconnor2013real-time}, SoftLIF function \cite{hunsberger2015spiking}, noisy softplus function \cite{liu2016noisy}, complementary cumulative distribution function of Gaussian distribution \cite{esser2015backpropagation}. These methods require training with less common activation functions and are not well compatible with state-of-the-art ANN architectures.

The other kind of conversion algorithms takes the advantage of the non-negativity of ReLU activation to approximate average firing rate. It was first introduced by Cao et al. in \cite{cao2015spiking}. Following work introduced methods such as data-based normalisation \cite{diehl2015fast} and dynamic threshold balancing \cite{Sengupta2019going} to improve the performance after conversion.

In \cite{neil2016learning,rueckauer2017conversion}, spiking versions of common ANN operations were introduced to convert modern ANN architectures and pre-trained ANN models. Hardware implementation was demonstrated by Esser et al. \cite{esser2016convolutional} with TrueNorth chip.

Although many SNN learning approaches have been presented, it is still very challenging to build a SNN with a large-scale deep architecture for obtaining high accuracy.

\section{Building Spiking ResNet}
The deep residual network (ResNet) is a variant of convolutional neural networks (CNNs) for the degradation problem of deep networks and has been widely used. Starting from the insight that a deep network constructed by adding identity mapping layers will not perform worse than original shallow network, He et al. \cite{he2016deep} let the stacked nonlinear layers approximate the mapping of $\mathcal{F}(x) := \mathcal{H}(x) - x$, where $\mathcal{H}(x)$ is the desired underlying mapping. Then the original mapping becomes a residual mapping: $\mathcal{H}(x) = \mathcal{F}(x) + x$. They hypothesised that residual mapping is easier to be optimised with current solvers and verified their hypothesis with empirical evidence. For the powerful capability of ResNet, this paper develops an effective conversion approach to build Spiking ResNets from trained ResNets.

\subsection{Spiking ResNet: An Overview}
In a word, conversion shapes an ANN into its biologically plausible counterpart, i.e., the SNN. Differences between ANN and SNN make the conversion not an easy task. Within an ANN, a neuron receives real-valued inputs from its previous neurons while a neuron within a SNN receives spike trains (series of spikes) as inputs from its previous neurons. Furthermore, neurons in an ANN process information by applying activation function to the summed inputs, whereas neurons in a SNN process information by integrating incoming spikes. Each incoming spike will induce a change to a neuron's postsynaptic potential (PSP). When PSP reaches certain threshold, the neuron fires a spike and passes information on to its next neurons.

\begin{figure}[ht!]
  \centering
          \resizebox{0.65\linewidth}{!}
        {
  \includegraphics[width=\linewidth]{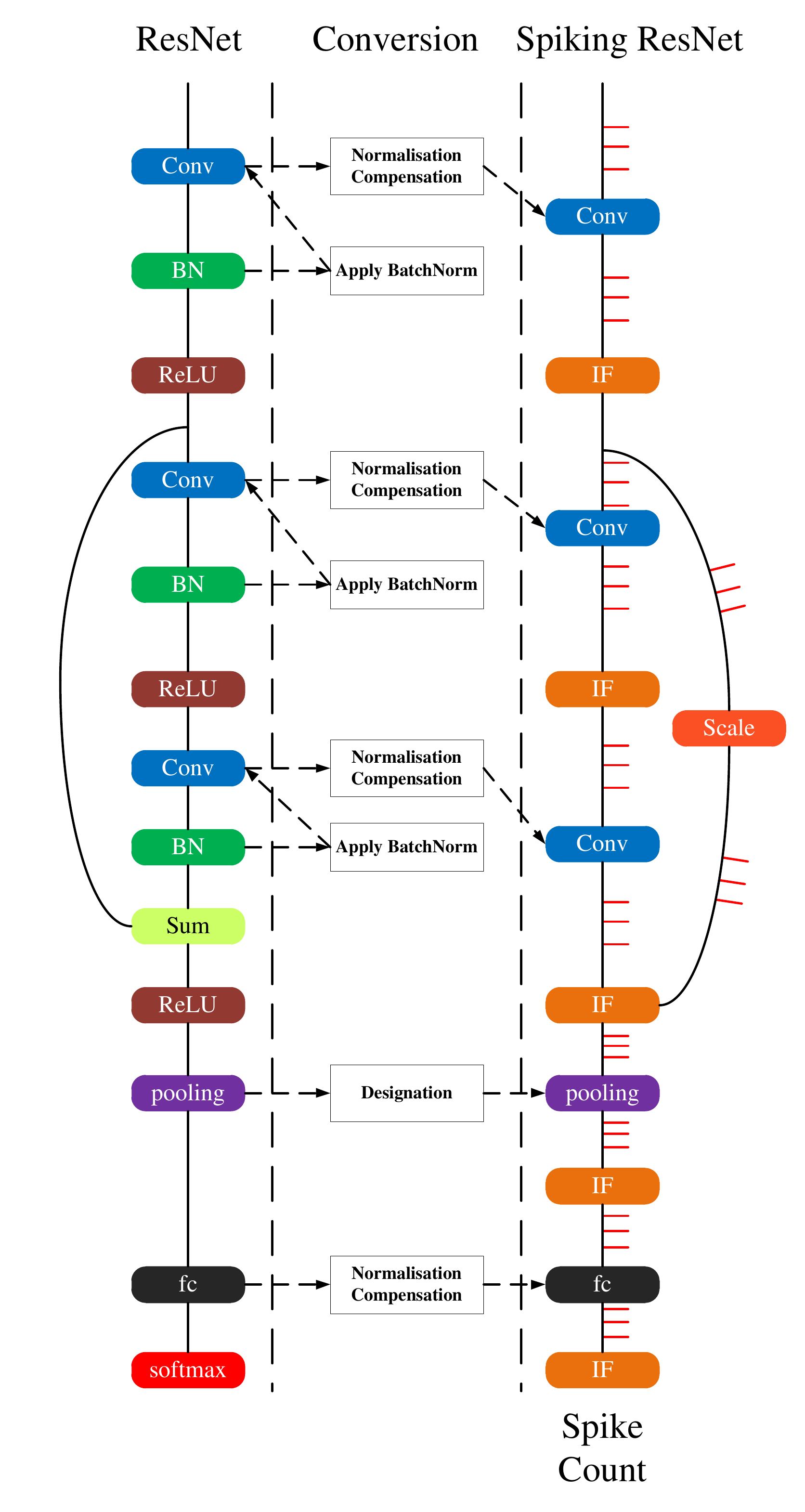}
  }
  \caption{Architectural overview of our conversion approach}\label{fig:arch}
\end{figure}

The conversion of networks comprised of two different types of neurons is based on the postulation that activation in ANN approximates firing rate of spiking neuron. Under this postulation, we map ANN to SNN by replacing artificial neuron with spiking neuron and creating synaptic connections between neurons. In Figure \ref{fig:arch}, we illustrate the architectural overview of the whole process. The activation function, ReLU layer, is replaced with integrate-and-fire (IF) neuron. Sum layer is no longer needed since IF neuron implicitly implements addition during the integration of spikes. Convolution layer is replaced with a layer of synaptic connections that resembles convolutional operation. Weights of convolution layer are mapped to corresponding synapse layer and biases are converted to constant current injected to spiking neurons \cite{rueckauer2017conversion}. Similarly, pooling and fully-connection are replaced by synapses that resemble these operations. For average pooling, synaptic weights are fixed to $1/PoolSize^2$. For max pooling, we use logical comparison to select spikes only from neuron with highest firing rate and inhibit other incoming spikes by setting synaptic weights to zero. An extra layer of IF neuron is added after pooling layer or fully-connected layer to integrate spikes from these types of synaptic connections. Since training is already finished before conversion, batch normalisation is directly applied to shape the convolution layer that precedes batch normalisation layer. Weights and biases in convolution layer are scaled to
\begin{equation}\label{eq:1}
  \widetilde{W} = \dfrac {\gamma W}{\sigma} , \quad \widetilde{b} = \dfrac{\gamma(b - \mu)}{\sigma} + \beta ,
\end{equation}
where $\mu$ is the mean, $\sigma$ is the variance, $\beta$ and $\gamma$ are two learnable parameters \cite{ioffe2015batch}.

Before we map weights in ANN to SNN, we must note that activations of ANN can be any positive real number as we employed ReLU function during training. Meanwhile, firing rate of spiking neuron is fixed in a region [0, $r_{max}$], where $r_{max}$ is the max firing rate as a biological neuron can fire no more than one spike within a brief interval (a single time step). For simplicity, we suppose neurons can fire as much as they could, i.e., firing a spike at each time step. In order to have activations in ANN matching firing rate in SNN (both in a region of [0, 1]), weights and biases in ANN are jointly normalised by

\begin{equation}\label{eq:2}
  \overline{W} = \dfrac{\lambda^{l-1}}{\lambda^l}W, \quad \overline{b} = \dfrac{1}{\lambda^l}b,
\end{equation}

where $\lambda^l$ is the max 99.9\% or 99.99\% activation at layer $l$ instead of the actual max activation as using the actual max activation is more prone to be susceptible to outliers \cite{rueckauer2017conversion}.

\subsection{Conversion Model of Residual Network}
\begin{figure*}[!htb]
  \centering
  \includegraphics[width=0.95\textwidth]{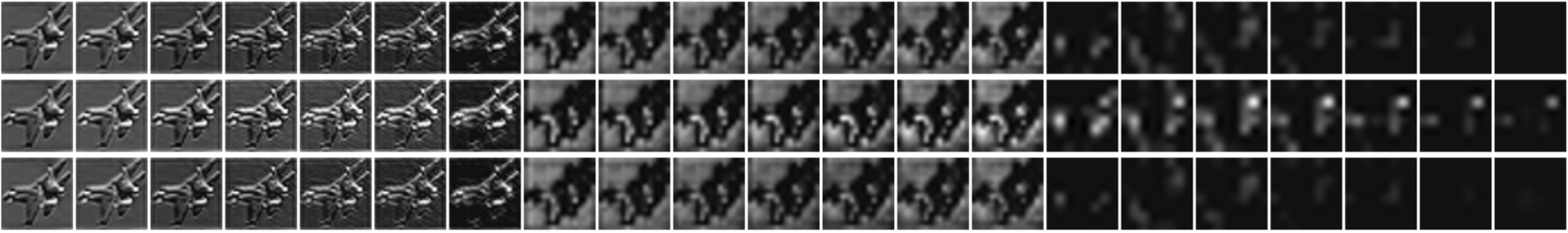}\\
  \caption{Illustrations of  feature maps of each residual block in ResNet-44 with an example input from CIFAR-10. The first row presents feature maps of original ANN. Second row presents feature maps of spiking ResNet-44 without its shortcuts normalised. Third row presents feature maps of spiking ResNet-44 with its shortcuts normalised.
  }\label{fig:fm}
\end{figure*}

Different from linear structures employed in previous methods, the topology of residual network is a directed acyclic graph. This makes normalisation methods in previous literature incompatible. When we convert residual network to its spiking version, new synaptic weights are introduced in shortcuts as SNN creates synaptic connections between any two neighboured layers of neurons. Although the original weights in shortcuts are set to 1 to perform identity mapping, those weights should also be scaled accordingly as weights in convolutional synapses being normalised. Otherwise, if we normalise stacked layers only, i.e., convolutional synapses, firing rates of spiking neurons would notably deviate from their corresponding activations. In Figure \ref{fig:fm}, we demonstrated an example of mismatch between firing rates and activations with the shortcuts unnormalised.

\begin{figure}[h]
  \centering

        \resizebox{0.65\linewidth}{!}
        {
            \includegraphics[width=\linewidth]{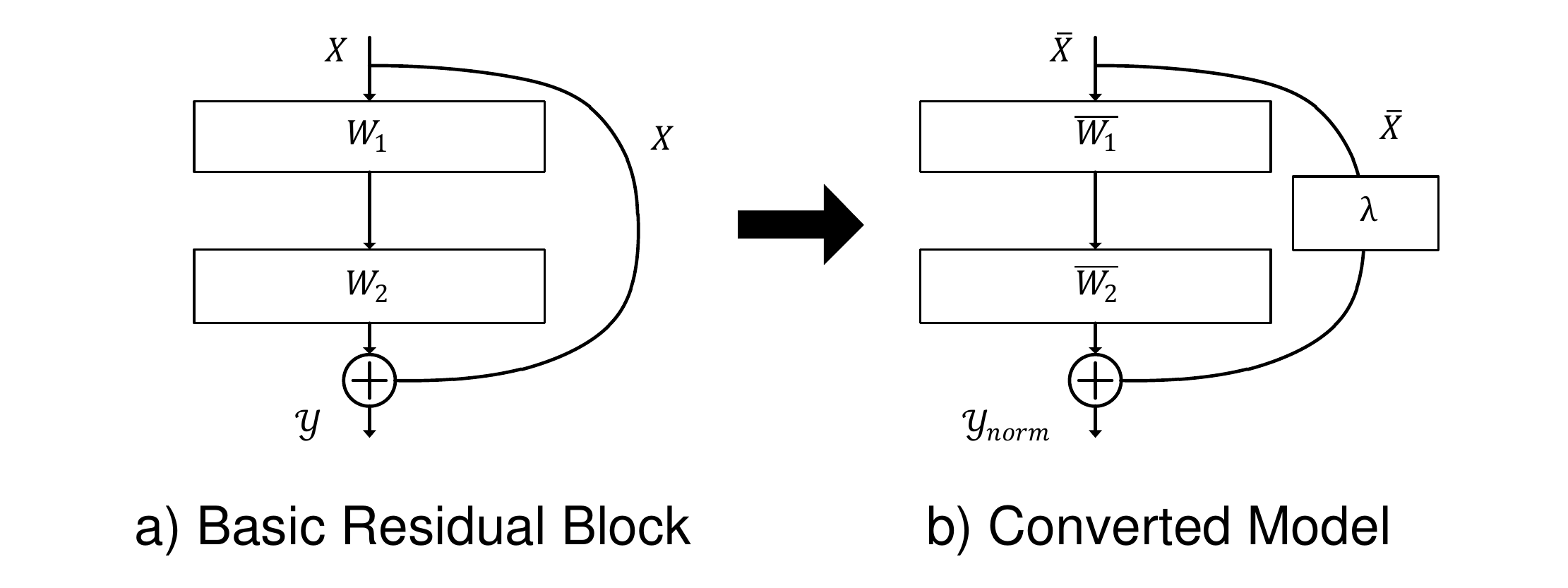}
        }

  \caption{Conversion model for a shortcut block}\label{fig:basic}
\end{figure}

To derive a good conversion model to scale shortcut, we consider the case of a basic residual block in ResNet, shown in Figure \ref{fig:basic} (a). Let $max_{in}$, $max_{int}$, $max_{out}$ denote max activations in input $X$, intermediate ReLU unit, output ReLU unit, respectively. $W_1$, $b_1$, $W_2$, $b_2$ are weights and biases in two weight layers. The original output of a basic residual block is determined by following equation:
\begin{equation}\label{eq:3}
  y = W_2(W_1X + b_1) + b_2 + X
\end{equation}

Then we let activations in ResNet match the range of firing rates in S-ResNet as spiking neurons are not allowed to fire multiple spikes at one time step. Otherwise, information carried by high activation values will be lost during the conversion. For the stacked layers, weights and biases at each layer are normalised by the maximum possible activation that is calculated from the training set. This leads to new weights, biases in the residual block and new input as well due to the change of weight in previous layers:

\begin{equation}\label{eq:4}
\begin{split}
  \overline{X} = \dfrac{1}{max_{in}}X, \quad \overline{W_1} = \dfrac{max_{in}}{max_{int}} W_1,\quad \overline{b_1} = \dfrac{1}{max_{int}}b_1,  \\
  \overline{W_2} = \dfrac{max_{int}}{max_{out}} W_2, \quad \overline{b_2} = \dfrac{1}{max_{out}}b_2.
\end{split}
\end{equation}

With new weights, biases and input, the output of the block is updated to
\begin{equation}\label{eq:5}
\begin{split}
y'    & = \overline{W_2}(\overline{W_1}\overline{X} + \overline{b_1}) + \overline{b_2} +  \overline{X}\\
            & =  \dfrac{W_2(W_1x + b_1) + b_2}{max_{out}} + \dfrac{1}{max_{in}}X \\
\end{split}
\end{equation}

However, in accordance to the basic idea of normalisation, no matter how the weights, biases or input change, the normalised output of one layer must always be the original activation normalised by its maximum possible activation:
\begin{equation}\label{eq:6}
  y_{norm} = \dfrac{y}{max_{out}} = \dfrac{W_2(W_1X + b_1) + b_2 + X}{max_{out}}
\end{equation}

By comparing Eq.(\ref{eq:6}) with Eq.(\ref{eq:5}),  we find out that the last term in Eq.(\ref{eq:5}), i.e., the shortcut connection should be scaled with a factor $\lambda$ (Figure \ref{fig:basic} (b)), where $\lambda = \dfrac{max_{in}}{max_{out}}$. Therefore, in spiking residual network, synaptic weights in shortcut should be scaled to $\dfrac{max_{in}}{max_{out}}$.

\subsection{Compensation of Propagation Error}
The conversion method discussed in this paper is based on the one-to-one correspondence between artificial neuron and spiking neuron: the normalised activation of ReLU is approximated by the firing rate of spiking neuron. For an artificial neuron $i$, its ReLU activation is computed as:
\begin{equation}\label{eq:7}
  a_i = \sum\limits^{\Xi}_{j=1} W_{ij}a_{j} + b_{i} \ ,
\end{equation}

where $a_i \in [0,1], \ $$\Xi$ is the set of all neurons feed their outputs to neuron $i$, neuron $j \in \Xi$, $W$ is weight and $b$ is bias.

After conversion, weights and biases of artificial neurons are directly mapped to IF neurons that resets its membrane potential by subtraction (neuron model used in this paper). If the intervals between spikes are larger than the refractory period, $N_i(t)$, i.e., the number of spikes neuron $i$ has fired from the start of simulation to time step $t$ satisfies

\begin{equation}\label{eq:8}
  N_{i}\left( t\right) = \dfrac{\sum\limits^{\Xi}_{j=1} W_{ij}N_{j}\left( t\right)+b_{i}t - V_i\left( t\right)}{\theta} \ ,
\end{equation}

where $\theta$ is the firing threshold, $V_i\left( t\right)$ is the membrane potential at the time $t$ and $\Xi$ is the set of neurons connected to neuron $i$ through pre-synaptic connections. For simplicity, we set $\theta$ to 1. The average firing rate of neuron $i$ is determined by:
\begin{equation}\label{eq:9}
  r_{i}\left( t\right) = \dfrac{N_{i}\left( t\right)}{t} = \sum\limits^{\Xi}_{j=1} W_{ij}r_{j}\left( t\right)+b_{i} - \dfrac{V_i\left( t\right)}{t}
\end{equation}

As is seen from the difference between Eq.(\ref{eq:7}) and Eq.(\ref{eq:9}), the approximation of ReLU activation incurs an amount of deviation at each neuron by the last term in Eq.(\ref{eq:9}), i.e., the membrane potential that is not yet integrated as spikes. Here, we define the amount of deviation $\dfrac{V_i\left( t\right)}{t}$ as propagation error $E_{i}(t)$. To explicitly determine the relation between ReLU activation and firing rate, we expand the recursive Eq. \ref{eq:9} iteratively with the firing rates of each pre-synaptic neuron:
\begin{equation}\label{eq:10}
\begin{split}
  r_{i}\left( t\right) = & a_{i} - \Bigg\{ E_{i}\left( t\right) + \sum\limits^{\Xi}_{j=1} W_{ij}E_{j}\left( t\right)   \\
                             & + \sum\limits^{\Theta}_{j=1} W_{ij}\sum\limits^{\Theta_1}_{j_1=1} W_{jj_1} \ldots \sum\limits^{\Xi_k}_{j_k=1} W_{j_{k-1}j_{k}}E_{j_{k}}\left( t\right) \\
                             & + \sum\limits^{\Theta}_{j=1} W_{ij} \ldots \sum\limits^{\Theta_k}_{j_k=1} W_{j_{k-1}j_{k}} \ldots \sum\limits^{\Xi_n}_{j_n=1} W^{2}_{j_{n - 1}j_{n}}E_{j_{n}}\left( t\right) \Bigg\} \\
\end{split}
\end{equation}

We let $\Omega$ denote the set of spiking neurons that directly receive input impulses. $\Theta$ denotes the set of neurons in $\Xi$ but not in $\Omega$ ($\Theta \subseteq \Xi, \Theta \cap \Omega = \emptyset$). $\Xi_k$ denotes the set of neurons connected to neurons in $\Theta_{k-1}$ and we have $\Xi_{n} \subseteq  \Omega$, i.e., $\Theta_{n} = \emptyset$. From Eq.\ref{eq:10}, we know that the propagation error accumulates as information propagates from pre-synaptic neurons to post-synaptic neurons. Let $\overline{E_i}$ denote the collection of all accumulated propagation errors in Eq. \ref{eq:10} (terms within the braces). If a SNN is constructed with a stack of feed-forward layers, Eq.\ref{eq:10} indicates that $\overline{E_i}$ grows as network goes deep. In this case, propagation error is a major factor that obstructs the building of large-scale spiking network as firing rates $r_{i}\left( t\right)$ at deep layers no longer approximate ReLU activations:
\begin{equation}\label{eq:11}
  r_{i}\left( t\right) = a_{i} - \overline{E_i}
\end{equation}

Intuitively, we can enlarge the firing rates to match the ReLU activation if we know exactly the total amount of propagation error $\overline{E_i}$ (Eq. \ref{eq:11}). However, it is very difficult to solve this problem for its complexity and non-determinacy.

\begin{figure}[htb!]
    \centering
    \begin{subfigure}[b]{0.9\linewidth}
        \centering
        \resizebox{0.6\linewidth}{!}
        {
            \includegraphics[width=4in]{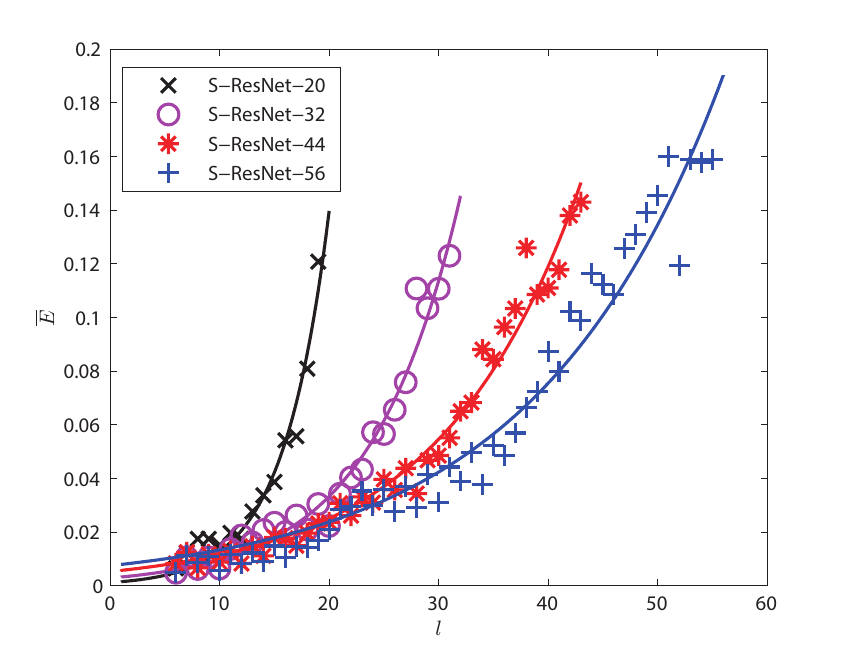}
        }
        \caption{CIFAR-10 with S-ResNets of depth 20, 32, 44, 56}
        \label{fig:P1}
    \end{subfigure}

    \begin{subfigure}[b]{0.9\linewidth}
        \centering
        \resizebox{0.6\linewidth}{!}
        {
            \includegraphics[width=4in]{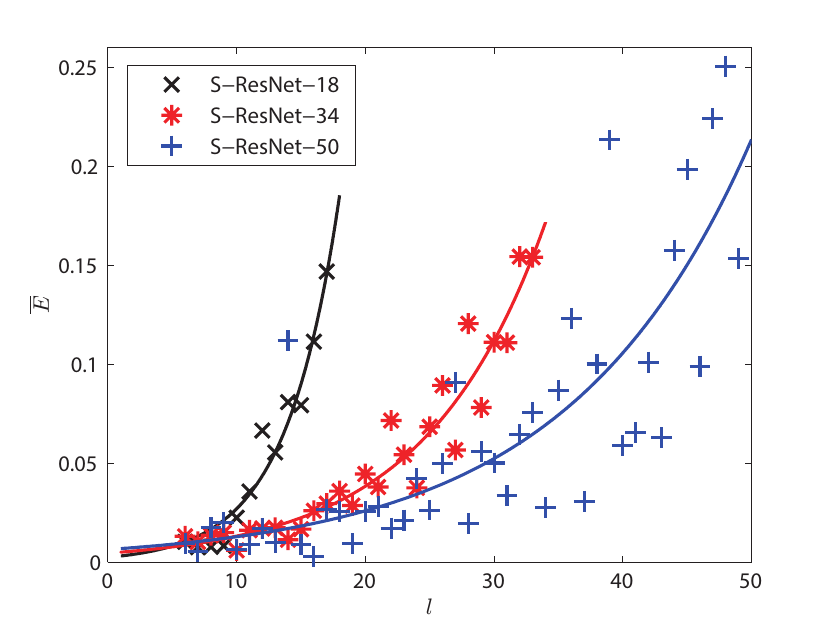}
        }
        \caption{ImageNet with S-ResNets of depth 18, 34, 50}
        \label{fig:P2}
    \end{subfigure}

    \caption{Growth of $\overline{E_i}$ on CIFAR-10 and ImageNet}
    \label{fig:fr}
\end{figure}

\begin{table}[htb!]
\centering
\begin{threeparttable}
\caption{Parameters and Goodness of Fit}\label{tab:goodness}
\begin{tabular}{ccccc}
\hline
\multirow{3}{*}{Dataset}     &\multirow{3}{*}{Depth}  &\multirow{3}{*}{p}    &\multirow{3}{*}{q}      &\multirow{3}{*}{$Adj. R^2$} \\
                            &&&&\\
                            &&&&\\\hline
\multirow{4}{*}{CIFAR-10}   &20   &0.001168 &1.270  &0.9562   \\
                            &32   &0.002853 &1.131  &0.9717   \\
                            &44   &0.005269 &1.081  &0.9753   \\
                            &56   &0.007478 &1.059  &0.9602   \\\hline
\multirow{3}{*}{ImageNet}   &18   &0.002418 &1.273  &0.9498   \\
                            &34   &0.040488 &1.113  &0.9192   \\
                            &50   &0.006339 &1.073  &0.6598   \\\hline
\end{tabular}

\end{threeparttable}
\end{table}

Based on our experimental observation, a certain pattern exists in the growth of $\overline{E_i}$ with respect to network depth. Figure \ref{fig:fr} shows how $\overline{E_i}$ of max firing neuron at each layer changes as S-ResNet goes deeper. The data is collected after a simulation of 350 time steps and averaged across 50,000 training samples in CIFAR-10 and 50,000 validation samples in ImageNet, respectively. We try to simplify the problem by building an approximation model of $\overline{E_i}$. We assume that accumulated propagation error $\overline{E_i}$ is dependent upon network depth and establish a regression model to describe their relationship:
\begin{equation}\label{eq:12}
  \overline{E_i} = p \cdot q^l \ ,
\end{equation}
where $l$ is the network depth, $p$ and $q$ are two constants. Fitting parameters and goodness of fit are presented in Table \ref{tab:goodness}, indicating that our model is faithful for most cases.

With the established model of $\overline{E_i}$, we adjust the weights of neurons to let them accordingly overfire a little bit to counterbalance the growth of propagation error. This reduces the problem to the minimisation of the $\overline{E}$ (summation of each neuron's $\overline{E_i}$):

\begin{equation}\label{eq:13}
  \argmin_{W_i} \sum\limits^{\Phi}_{i=1} \overline{E_i} ( W_i, l, t)
\end{equation}
$\Phi$ is the set of all neurons in SNN. Considering the exponential model of $\overline{E_i}$, we use a simple weight update rule to reduce the number of parameters instead of fiddling each layer's weights:
\begin{equation}\label{eq:14}
  W' = W \cdot \zeta,
\end{equation}
where $\zeta$ is the compensation factor. With a subset of training data, $\zeta$ is searched step-wisely around $q$, bounded by region (1, $\tau_{max}$), where $\tau_{max}$ is the reciprocal of max firing rate at the last layer.

\section{Experiment}
For evaluation, we conducted experiments on four publicly available datasets: MNIST \cite{lecun1998gradient}, CIFAR-10, CIFAR-100 \cite{krizhevsky2009learning} and ImageNet 2012 \cite{ILSVRC15}. We employed residual networks with the same architectures as those defined in \cite{he2016deep}. All experiments were conducted with deep learning library MatConvNet \cite{vedaldi2015matconvnet}.

\subsection{Experiment on Shortcut Model}
\begin{table}[htb!]
\centering
\begin{threeparttable}
\caption{Error rates of S-ResNets with or without shortcut model on CIFAR-10}\label{tab:shortcut}
\begin{tabular}{c|c|cc|cc}
\hline
\multirow{3}{*}{Depth}     &\multirow{3}{*}{ResNet}       &\multicolumn{4}{c}{S-ResNet} \\\cline{3-6}
                           &                      &\multicolumn{2}{c|}{\specialcellc{without shortcut\\ model}}  &\multicolumn{2}{c}{\specialcellc{with shortcut \\ model}}    \\\cline{3-6}
                           &                      &err. &(acc. loss)   &err. &(acc. loss)    \\\hline
20                         &8.01\%   &10.55\%  &2.54\%  &8.21\% &0.20\% \\
32                         &7.34\%   &14.62\%  &7.28\%  &8.30\% &0.96\% \\
44                         &7.15\%   &15.44\%  &8.29\%  &\textbf{8.02}\% &0.87\% \\
56                         &7.01\%   &16.67\%  &9.66\%  &8.36\% &1.35\% \\
110                        &6.53\%   &20.77\%  &14.24\% &12.18\% &5.65\% \\\hline
\end{tabular}

\end{threeparttable}
\end{table}

Shortcut conversion model was tested with CIFAR-10. In CIFAR-10, there are 50,000 training images and 10,000 testing images, which belong to 10 classes. The residual networks were trained with the standard stochastic gradient descent (SGD) with a momentum of 0.9. Training data is augmented by padding 4 pixels on each side of the image and randomly sampled from the padded image or its horizontal flip with a $32 \times 32$ crop \cite{he2016deep}. After training, we apply our approach to build the spiking residual network. Normalisation factors are obtained by surveying activations from the whole training set.

We trained the residual networks of depth 20, 32, 44, 56, 110 and build their spiking counterparts with or without shortcut normalisation applied. The results are summarised in Table \ref{tab:shortcut}. Across all depths, SNNs with their shortcuts normalised achieved much lower error rates than those without shortcut normalisation. At depth 20, 32, 44, conversion with shortcut normalisation notably achieved less than 1\% loss of classification error rates, with only 1.35\% error loss for the depth 56. Note that the error compensation has not been applied yet.

\subsection{Experiment on Error Compensation }
To assess effectiveness of the error compensation, we applied error compensation to both residual networks and classical convolutional neural networks. The classical convolutional network, called "plain CNN", is a stack of convolutional layers without shortcut connections, which we implemented it as a VGG-like network \cite{simonyan2014very} whose depth is exactly the same as its residual counterpart.

On CIFAR-10, networks were trained under the same protocol described in Section 4.1. Experimental results shown in Table \ref{tab:comp} demonstrate that error compensation consistently improve the performance and reduce the propagation error of both the ResNet and plain CNN over all depth settings (propagation error $\overline{E}$ in Table \ref{tab:comp} and \ref{tab:imagenet} is averaged by total number of neurons within network). The improvements are more obvious in deeper network as deeper network incurs greater propagation error. We also note that improvements on plain CNN are much higher than those on residual network, which indicates that conversion process induces more errors in plain network than in residual network. This suggests that difference in network architecture affects the growth rate of propagation error and has critical impact on the final performance of converted SNNs.

\begin{table*}[htb!]
\centering
\begin{threeparttable}
\caption{Experimental results (error rates) with compensation on CIFAR-10}\label{tab:comp}
\begin{tabular}{c|c|cc|cc}
\hline
\multirow{2}{*}{Depth}     &\multicolumn{5}{c}{Plain CNN}     \\\cline{2-6}
                           &CNN      &S-CNN  &$\overline{E}$  &\specialcellc{S-CNN \\(comp.)}  &$\overline{E}$ \\\hline
20                         &9.30\%   &11.26\% &0.006068 &10.60\% &0.005076           \\
32                         &10.15\%  &15.82\% &0.009681 &13.88\% &0.007916           \\
44                         &10.48\%  &38.64\% &0.014039 &30.88\% &0.011564           \\
56                         &12.47\%  &53.28\% &0.015086 &41.49\% &0.011438           \\\hline
\multirow{2}{*}{Depth}     &\multicolumn{5}{c}{Residual Network}\\\cline{2-6}
                           &ResNet &S-ResNet &$\overline{E}$    &\specialcellc{S-ResNet\\(comp.)} &$\overline{E}$\\\hline
20                         &8.01\% &8.21\% &0.004226 &8.18\% &0.004068  \\
34                         &7.34\% &8.30\% &0.005383 &7.96\% &0.004772  \\
44                         &7.15\% &8.02\% &0.006351 &7.63\% &0.005476  \\
56                         &7.01\% &8.36\% &0.007927 &7.67\% &0.006599  \\\hline

\end{tabular}
\end{threeparttable}
\end{table*}

\begin{table*}[htb!]
\centering

\resizebox{\textwidth}{!}{
\begin{threeparttable}
\caption{Experimental results (error rates) on ImageNet}\label{tab:imagenet}
\begin{tabular}{cc|ccc|ccc}
\hline
                    \multirow{2}{*}{Depth}         &ResNet            &\multicolumn{3}{c|}{S-ResNet}   &\multicolumn{3}{c}{\specialcellc{S-ResNet (comp.)}}  \\
                                                   &Top1-Error     &Top1-Error &(acc. loss) &$\overline{E}$ &Top-1 Error  &(acc. loss) &$\overline{E}$\\\hline
                        18                         &30.79\%       &31.65\%   &0.86\%  &0.002545  &31.64\%     &0.84\%       &0.002481        \\
                        34                         &27.12\%       &28.49\%   &1.37\%  &0.003657  &28.39\%     &1.27\%       &0.003253        \\
                        50                         &24.55\%       &27.69\%   &3.13\%  &0.004883  &27.25\%     &2.69\%       &0.004438        \\\hline

                                                   &Top-5 Error    &Top-5 Error&(acc. loss) & &Top-5 Error  &(acc. loss) & \\\hline
                        18                         &11.33\%       &11.76\%   &0.42\%  & &11.73\%     &0.39\%        &       \\
                        34                         &9.20\%        &9.86\%    &0.66\%  & &9.78\%      &0.58\%        &       \\
                        50                         &7.50\%        &9.34\%    &1.84\%  & &9.03\%      &1.52\%        &       \\\hline

\end{tabular}

\end{threeparttable}
}
\end{table*}

Furthermore, we evaluate the proposed S-ResNet with ImageNet 2012 dataset \cite{ILSVRC15}. The dataset comprises 1.2 million training images, 50,000 validation images and 100,000 test images. Residual networks used here are directly adopted from \cite{imageneres}. The spiking residual network is then validated on whole validation set of 50,000 images. Each image is resized to 224 by its short edge and then cropped into its $224 \times 224$ centre patch. Table \ref{tab:imagenet} summarises the results. Our approach achieved the top-1 error rate of 27.25\% (at the depth 50, with the error loss of 2.69\%) and the top-5 error rate of 9.03\% (at the depth 50, with the error loss of 1.52\%). Both of them are the best performance on ImageNet 2012, compared with the reported in the literature. The results demonstrate that error compensation consistently improves the performance of both top-1 and top5 error rates at different depths.

\begin{table*}[htb!]
\centering
\begin{threeparttable}
\caption{Comparison with other conversion methods on MINIST, CIFAR-10, CIFAR100 and ImageNet.}\label{tab:best}
\begin{tabular}{lllll}
\hline
\multirow{2}{*}{Dataset}                 &\multirow{2}{*}{Depth}    &\multirow{2}{*}{ANN}       &\multirow{2}{*}{Converted SNN}        &\multirow{2}{*}{Increment}\\
                                         &&&&\\\hline

MNIST\cite{diehl2015fast}                &3        &0.86\%   &0.88\%    &0.02\% \\
MNIST\cite{rueckauer2017conversion}      &4        &0.56\%   &0.56\%    &0\%    \\
\textbf{MNIST(ours)}                     &8        &\textbf{0.41\%}   &\textbf{0.41\%}  &\textbf{0\%}\\\hline
CIFAR-10\cite{cao2015spiking}            &5        &20.88\%   &22.57\%    &1.69\% \\
CIFAR-10\cite{hunsberger2015spiking}     &5        &14.03\%   &16.46\%    &2.43\%\\
CIFAR-10\cite{esser2016convolutional}    &16       &N/A       &10.68\%    &N/A\\
CIFAR-10\cite{rueckauer2017conversion}   &9        &8.09\%   &9.15\%    &1.06\%\\
CIFAR-10-VGG \cite{Sengupta2019going}    &16       &8.3\%    &8.45\%    &0.15\%\\
CIFAR-10-ResNet \cite{Sengupta2019going} &20       &10.9\%   &12.54\%   &1.64\%\\

\textbf{CIFAR-10(ours)}                  &44       &\textbf{7.15\%}   &\textbf{7.63\%}  &\textbf{0.48\%}\\
CIFAR-100\cite{esser2016convolutional}   &16       &N/A       &34.52\%    &N/A \\
\textbf{CIFAR-100(ours)}                 &44       &\textbf{29.82\%}   &\textbf{31.44\%}  &\textbf{1.62\%}\\\hline

ImageNet-VGG \cite{Sengupta2019going}    &16       &29.48 \%\ (10.61\%) &30.04 \%\ (10.99\%)    &0.56\%\ (0.38\%)\\
ImageNet-ResNet \cite{Sengupta2019going} &34       &29.31\% \ (10.31\%)   &34.53\% \ (13.67\%)    &5.22\% \ (3.36\%) \\

\textbf{ImageNet (ours)}                        &34        &\textbf{27.12\% \ (9.20\%)} &\textbf{28.39\% \ (9.78\%)}       &\textbf{1.27\% \ (0.58\%)}\\
\textbf{ImageNet (ours)}                        &50        &\textbf{24.55\% \ (7.50\%)} &\textbf{27.25\% \ (9.03\%)}       &\textbf{2.7\% \ (1.53\%)}\\\hline

\end{tabular}
\footnote{a} top-5 errors on ImageNet are included within the round brackets
\end{threeparttable}
\end{table*}

\subsection{Comparison with Other Deep SNNs}
In Table \ref{tab:best}, we present a summary of our results achieved on MNIST, CIFAR-10, CIFAR-100, ImageNet along with comparison with the performance of other deep SNNs. By and large, our approaches allow us to achieve better performance with deeper networks than state-of-the-art SNNs. On MNIST, we achieved loss-less performance with a relative shallow network (depth of 8). On CIFAR-10 we achieved an error rate of 7.63\% with Spiking ResNet-44 (only error loss of 0.48\%), and obviously outperformed the other SNN methods, for example, those in \cite{hunsberger2015spiking,cao2015spiking,Sengupta2019going,rueckauer2017conversion,esser2016convolutional}

On CIFAR-100, we performed exactly the same protocol as we did on CIFAR-10, yielding an error rate of 31.44\% with Spiking ResNet-44, the state of the art performance. On ImageNet, we achieved 27.25\% top-1 error, defeated the SNN approaches in the literature. Overall, the proposed Spiking ResNet achieves the best performance on CIFAR-10, CIFAR-100 and ImageNet 2012, compared with the state-of-the-art SNN approaches. It is the first time to build a SNN deeper than 40 on a large-scale dataset while keeping comparable performance to ANN.

\subsection{Analysis of Energy Consumption}
SNN's potential to achieve revolutionary energy-efficiency with emerging neuromorphic hardware is one of its advantages over traditional computing platform. We estimate power consumption for both ResNet and S-ResNet, assuming that they are run on their advanced hardware.  For ResNet, we employ FPGA of Intel Stratix 10 TX \cite{intelstratix10} for estimation. Intel Stratix 10 TX was fabricated using 14nm technology and released in 2018. It is considered as one of the most powerful and energy-efficient platforms to date among ready-made FPGAs. Its power efficiency is up to 80 GFLOPS/Watt, i.e., it operates at a cost of 12.5pJ per FLOP. For S-ResNet, we employ the neuromorphic chip of ROLLS \cite{ning2015reconfigurable} for estimation. It was fabricated using a 180nm CMOS process in 2015. As reported in reported in \cite{indiveri2015neuromorphic}, the ROLLS consumes 77fJ per SOP (synaptic operation \cite{merolla2014million}).

The ImageNet dataset is adopted to estimate the amount of energy required to classify one single image from the dataset. We first determine the number of operations (FLOP for ResNet and SOP for S-ResNet) required for the task and then multiple it with platform's power efficiency for final result.The number of operations goes as follows: 1.82 GFLOP, 3.67 GFLOP, and 4.12 GFLOP respectively for ResNet-16, ResNet-34, and ResNet-50. 33.13 GSOP, 65.28 GSOP and 78.29 GSOP respectively for S-ResNet-16, S-ResNet-34, and S-ResNet-50. With reference to each platform's energy efficiency, the power consumption of ResNet is more than 9 times greater than energy consumed by S-ResNet across all depths. This result has shown us promising future of SNN-based neuromorphic systems as the estimated power consumption of SNNs implemented with neuromorphic hardware beats that of ANNs implemented with one of the top energy-efficient FPGA platforms by a degree of magnitude. Note that the FPGA Intel Stratix 10 TX is manufactured using 14 nm technology while the ROLLS is with only 180nm technology, six generations APART from 14nm technology.

\section{Conclusion}
It is still very challenging to build a deep spiking neural networks in the neuromorphic computing community. In this paper, we presented an efficient approach for converting a deep residual network to its spiking counterpart, which can be finally moved to energy-efficient neuromorphic hardwares. The proposed conversion model enables us to improve the performance of SNNs by employing deeper structure. Furthermore, we designed a compensation mechanism to reduce the discretisation error accumulated by going deeper of a SNN. Compared with the state-of-the-art SNN approaches, our proposed Spiking ResNet achieved the best performance on CIFAR-10, CIFAR-100 and ImageNet 2012. The proposed approach has potential to be extended for other deep ANN architectures.

\bibliographystyle{IEEEtran}
\bibliography{Alruna}

\end{document}